%% file: neurips_2025.tex
\documentclass{article}

\usepackage[preprint]{neurips_2025}

\usepackage[utf8]{inputenc} 
\usepackage[T1]{fontenc}    
\usepackage{url}            
\usepackage{booktabs}       
\usepackage{amsfonts}       
\usepackage{nicefrac}       
\usepackage{microtype}      
\usepackage{booktabs}
\usepackage{pifont}
\usepackage{multirow}      
\usepackage{graphicx}      
\usepackage{array}
\usepackage{subcaption}
\usepackage[table]{xcolor}
\usepackage{authblk}
\usepackage{enumitem}
\newcommand{\cmark}{\textcolor{green!60!black}{\ding{51}}}
\newcommand{\xmark}{\textcolor{red}{\ding{55}}}

\definecolor{SelfColor}{rgb}{0.913,0.443,0.196}
\definecolor{UrlColor}{rgb}{0.7098,0.009,0.0}
\definecolor{RefColor}{rgb}{0.082,0.376,0.510}

\usepackage[colorlinks,
            linkcolor = UrlColor,
            urlcolor  = UrlColor, 
            citecolor= RefColor,
]{hyperref}
\usepackage[capitalize,noabbrev]{cleveref}

\title{S\textsc{urprise}3D: A Dataset for Spatial Understanding and Reasoning in Complex 3D Scenes}

\author{%
    \textbf{Jiaxin Huang}\textsuperscript{1\dag},
    \textbf{Ziwen Li}\textsuperscript{1\dag},
    \textbf{Hanlue Zhang}\textsuperscript{1\dag},
    \textbf{Runnan Chen}\textsuperscript{2},
    \textbf{Zhengqing Gao}\textsuperscript{1},\\
    \textbf{Xiao He}\textsuperscript{3},
    \textbf{Yandong Guo}\textsuperscript{3},
    \textbf{Wenping Wang}\textsuperscript{4},
    \textbf{Tongliang Liu}\textsuperscript{1,2\ddag},
    \textbf{Mingming Gong}\textsuperscript{1,5\ddag}
}

\affil
{
  \textsuperscript{1}MBZUAI, 
  \textsuperscript{2}The University of Sydney, 
  \textsuperscript{3}AI2Robotic \\
  \textsuperscript{4}Texas A\&M University,
  \textsuperscript{5}The University of Melbourne
}

\begin{document}

\maketitle

\input{sec/0_abstract}
\input{sec/1_introduction}

\input{sec/2_related_works}

\input{sec/3_benchmark}
\input{sec/4_experiment}

\bibliography{bibliography}
\bibliographystyle{plain}

\end{document}

%% file: sec/0_abstract.tex
\begin{abstract}
  The integration of language and 3D perception is critical for embodied AI and robotic systems to perceive, understand, and interact with the physical world. Spatial reasoning, a key capability for understanding spatial relationships between objects, remains underexplored in current 3D vision-language research. Existing datasets often mix semantic cues (e.g., object name) with spatial context, leading models to rely on superficial shortcuts rather than genuinely interpreting spatial relationships. To address this gap, we introduce S\textsc{urprise}3D, a novel dataset designed to evaluate language-guided spatial reasoning segmentation in complex 3D scenes. S\textsc{urprise}3D consists of more than 200k vision language pairs across 900+ detailed indoor scenes from ScanNet++ v2, including more than 2.8k unique object classes. The dataset contains 89k+ human-annotated spatial queries deliberately crafted without object name, thereby mitigating shortcut biases in spatial understanding. These queries comprehensively cover various spatial reasoning skills, such as relative position, narrative perspective, parametric perspective, and absolute distance reasoning. Initial benchmarks demonstrate significant challenges for current state-of-the-art expert 3D visual grounding methods and 3D-LLMs, underscoring the necessity of our dataset and the accompanying 3D Spatial Reasoning Segmentation (3D-SRS) benchmark suite. S\textsc{urprise}3D and 3D-SRS aim to facilitate advancements in spatially aware AI, paving the way for effective embodied interaction and robotic planning. The code and datasets can be found in \href{https://github.com/liziwennba/SUPRISE}{https://github.com/liziwennba/SUPRISE}.
\end{abstract}

%% file: sec/1_introduction.tex
\section{Introduction}
Spatial reasoning lies at the heart of embodied AI and robotic systems~\cite{azzolini2025cosmos}. For agents to navigate real-world environments, manipulate objects, or interact meaningfully with humans, they must interpret instructions that are deeply rooted in 3D spatial layouts and context. Consider a robot waiter tasked with serving drinks to \textit{the table to the left of the couch}, or a household assistant that infers from \textit{I am thirsty} the intention of regaining the nearest cup. In both scenarios, the agent must go beyond identifying object categories, reasoning about spatial relationships, viewpoint-dependent references, and pragmatic intent. This type of language-guided spatial understanding is critical for tasks such as navigation, manipulation, and human-robot interaction~\cite{ma20243dsrbench}.


However, despite its importance, existing 3D vision language grounding benchmarks do not faithfully capture or evaluate spatial reasoning. Most prior datasets rely heavily on explicit object references, allowing models to localize targets by matching named categories or rigid templates, without requiring genuine spatial inference. As a result, current models achieve strong performance not by understanding spatial context but by exploiting superficial textual patterns or semantic shortcuts~\cite{zhu2024scanreason}.

We identify three major limitations of existing 3D vision-language benchmarks:

(1) \textbf{Overreliance on explicit queries}. Datasets such as ScanRefer~\cite{chen2020scanrefer} and ReferIt3D~\cite{referit3d} provide queries that including a object name (e.g., chair). Although challenging before the emergence of large language models (LLMs), such datasets have become increasingly solvable through category detection, often requiring little or no reasoning.

(2) \textbf{Limited and shallow reasoning coverage}. Recent efforts such as Intent3D~\cite{Intent_3D}, ScanReason~\cite{zhu2024scanreason}, Reason3D~\cite{reason3d} and Instruct3D~\cite{segpoint} have taken steps towards incorporating implicit queries and common sense. However, these benchmarks remain narrow in scope—ScanReason, for instance, contains only ~10K Q\&A pairs across five loosely defined reasoning types, and does not provide a rigorous or fine-grained definition of spatial reasoning in 3D contexts.

(3) \textbf{Template-driven or trivial spatial queries}. Many datasets~\cite{3dllm} formulate spatial queries using basic patterns (e.g., 'the object to the left') that do not require understanding geometric variability or occlusion and can often be resolved using positional heuristics.

\begin{table}[!t]
    \centering
    \small
    \caption{Comparison of major 3D vision-language datasets. 'Multi-target' indicates if a query refers to multiple objects. 'Lang Source' denotes whether language queries are human-annotated or generated via template or LLM. 'Shortcut Free' indicates if object names are avoided in queries.}
    \label{tab:datasets_revised}
    \renewcommand{\arraystretch}{1.2}
    \resizebox{0.9\linewidth}{!}{%
    \begin{tabular}{lcccccc}
    \toprule
    \textbf{Dataset} & \textbf{Output} & \textbf{Multi-target} & \textbf{Lang. Source} & \textbf{Spatial Reason} & \textbf{1st observ.} & \textbf{Shortcut Free} \\
    \midrule
    CLEVR3D~\cite{CLEVR3D}            & Lang            & -               & Template           & \xmark                & -                      & -               \\
    Scan2Cap~\cite{Scan2Cap}          & Lang            & -               & Human              & \xmark                & -                      & -               \\
    ScanQA~\cite{ScanQA}              & Lang            & \xmark               & Human              & \xmark                & -                      & -               \\
    3DVQA~\cite{etesam20223dvqa}      & Lang            & \xmark               & Human              & \xmark                & -                      & -               \\
    SQA3D~\cite{ma2022sqa3d}          & Lang            & -               & Human              & \cmark                & \cmark                 & \xmark          \\
    ScanScribe~\cite{3dvista}         & Lang            & \xmark               & Template,LLM       & -                     & -                      & -               \\
    3DMV-VQA~\cite{hong20233d}
                                      & Lang            & -               & Template           & \xmark                & \xmark                 & \xmark          \\
    M3DBench~\cite{li2024m3dbench}    & Lang            & \cmark               & LLM                & \xmark                & \xmark                 & \xmark               \\
    SceneVerse~\cite{jia2024sceneverse}
                                      & Lang            & \xmark               & Template,LLM       & \xmark                & \xmark                 & -               \\
    MSQA~\cite{linghu2024multi}       & Lang            & -               & Human,LLM          & \cmark                & \cmark                 & \xmark          \\
    VLA-3D~\cite{zhang2024vla03d0}    & Lang            & \cmark               & Template           & \cmark                & \cmark                 & \xmark          \\
    ExCap3D~\cite{yeshwanth2025excap3d0}
                                      & Lang            & \cmark               & LLM                & \xmark                & \cmark                 & \xmark               \\
    \midrule
    ReferIt3D~\cite{referit3d}        & BBox         & \cmark               & Human,Template     & \xmark                & \cmark                 & \xmark               \\
    ScanRefer~\cite{chen2020scanrefer}
                                      & BBox         & \xmark               & Human              & \xmark                & \cmark                 & \xmark               \\
    3D-DenseOG~\cite{huang2023dense}  & BBox         & \cmark               & Human              & \xmark                & \cmark                 & \xmark               \\
    Grounded 3D-LLM~\cite{grounded3DLLM}
                                      & Mask            & \xmark               & Template,LLM       & \cmark                & \xmark                 & \xmark          \\
    ScanEnts3D~\cite{abdelreheem2022scanents}
                                      & BBox            & \cmark               & Human              & \cmark                & \xmark                 & \xmark          \\
    PhraseRefer~\cite{yuan2022toward} & BBox              & \cmark               & Human              & \xmark                & \xmark                 & \xmark          \\
    EmbodiedScan~\cite{embodiedscan}  & BBox         & \cmark               & Template,LLM       & \cmark                & \cmark                 & \xmark          \\
    \midrule
    3D-LLM~\cite{3dllm}               & Lang + BBox     & \cmark               & LLM                & \cmark                & \xmark                 & \xmark          \\
    LL3DA~\cite{ll3da}                & Lang + BBox     & \cmark               & Template,LLM       & \xmark                & \cmark                 & \xmark          \\
    3DMIT~\cite{li20243dmit}         & Lang + BBox    & \xmark               & LLM                & \xmark                & \cmark                 & \xmark               \\
    3D-GRAND~\cite{yang2024_3D_GRAND}                          & Lang + BBox              & \cmark               & Template,LLM       & \xmark                & \xmark                 & \xmark          \\
    \midrule
    Segpoint~\cite{segpoint}                          & Mask            & \xmark               & LLM                & \cmark                & \xmark                 & \cmark          \\
    ScanReason~\cite{zhu2024scanreason}                          & BBox            & \xmark               & LLM                & \cmark                & \xmark                 & \cmark          \\
    Reason3D~\cite{reason3d}                          & Mask         & \xmark               & -                  & \cmark                & \xmark                 & \cmark          \\
    Intent3D~\cite{Intent_3D}                          & BBox         & \cmark               & LLM                & \xmark                & \xmark                 & \cmark          \\
    \midrule
    S\textsc{urprise}3D (Ours)
                                      & Mask         & \cmark               & Human,LLM & \cmark                & \cmark                 & \cmark          \\
    \bottomrule
    \end{tabular}}
\end{table}
These limitations lead to a recurring problem: models perform well by leveraging semantic priors and dataset biases rather than learning spatial reasoning. There is a critical need for a benchmark that disentangles spatial understanding from semantic recognition and poses queries that necessitate interpreting the scene geometry in context.
To address this gap, we introduce S\textsc{urprise}3D: a large-scale dataset and benchmark designed from the first principles to evaluate language-guided spatial reasoning in complex 3D scenes. Built on top of 900+ richly annotated indoor environments from ScanNet++ v2~\cite{scannetpp}, S\textsc{urprise}3D includes more than 200,000 query-object mask pairs, covering 2,800+ object classes. It is the first benchmark to support spatial reasoning segmentation at this scale, breadth, and level of annotation precision. Key features of S\textsc{urprise}3D include:

\textbf{Complex spatial queries}: Notably, we find that LLMs and MLLMs are incapable of generating spatial reasoning annotations with sufficient fidelity, necessitating a human-in-the-loop annotation process. 89K+ human-generated questions that require varied spatial reasoning. Queries investigate relative position recall (\textit{'the vase next to the left door'}), narrative perspective reasoning (describing objects from a moving observer's point of view), parametric perspective reasoning (specifying angles or offsets) and absolute distance reasoning (\textit{'the plamp 2 meters above the floor'}).

\textbf{Pragmatic language quality with Human Check}: All expressions follow Gricean conversational maxims (clarity, relevance, and brevity) and are vetted by multiple annotators to avoid ambiguity or bias. To resolve ambiguous references in cluttered environments, we adopt Gricean maxims to ensure clarity and informativeness, and introduce category-specific disambiguation rules that favor spatial, functional, or visual attributes depending on context.

\textbf{3D-SRS benchmark suite.} We introduce a formal evaluation framework tailored for 3D reasoning segmentation. It includes task definitions, metrics such as mask IoU and grounding precision, and diagnostic breakdowns between reasoning types. Our results demonstrate that state-of-the-art 3D vision-language models, when deprived of explicit naming, perform significantly worse, revealing their limitations in spatial understanding.

\textbf{Common sense and human intention reasoning.} In addition to spatial queries, we incorporate 110K LLM-generated questions that target pragmatic reasoning: for example, 'the object used to sit' (common sense) or 'the item someone might be reaching for' (intention). These complement spatial cues with functional and behavioral semantics, which are essential for embodied interaction.
 
By combining linguistic subtlety and geometric complexity, S\textsc{urprise}3D sets a new standard for 3D spatial intelligence benchmarks. Unlike existing datasets, our queries are implicit, ambiguous, and semantically lightweight: any correct object that satisfies the spatial constraint is valid, and object names are deliberately avoided. This forces models to rely on reasoning rather than recognition, and aligns with recent calls to evaluate deeper spatial understanding~\cite{zhu2024scanreason}.

Table~\ref{tab:datasets_revised} summarizes the key distinctions between S\textsc{urprise}3D and previous datasets. 
Our empirical evaluations further reveal that even the strongest existing 3D foundation models struggle under these conditions, highlighting a significant opportunity for innovation.
 Empirical evaluations with the state-of-the-art expert 3D visual grounding (VG) model and 3D-LLMs reveal a sharp drop in performance when deprived of semantic shortcuts, underscoring the unmet need for robust spatial reasoning abilities in current systems. Our main contributions are as follows.
\begin{itemize}[leftmargin=22pt,topsep=0pt, itemsep=1pt]
\item We introduce \textbf{S\textsc{urprise}3D}, a novel dataset of more than 200K language-guided 3D segmentation queries that cover spatial, common sense and human intention reasoning.
\item We define the \textbf{3D-SRS benchmark}, the first standardized protocol to evaluate spatial reasoning segmentation in 3D point clouds.
\item We show that current state-of-the-art 3D vision-language models underperform significantly on S\textsc{urprise}3D, indicating the need for new methods to handle implicit reasoning and spatial understanding in complex 3D scenes.
\end{itemize}

We hope S\textsc{urprise}3D will serve as a foundation for future research in spatially grounded 3D understanding and drive progress in embodied AI, robotics, and spatial intelligence.

%% file: sec/2_related_works.tex
\section{Related Works}
\subsection{Spatial Reasoning in 3D Vision-Language Model}
Understanding natural language in 3D scenes~\cite{huang2025mllm,chen2024ovgaussian,chen2023bridging,lu2023see,chen2022zero,chen2023clip2scene,chen2023towards,yang2024_3D_GRAND,jia2024sceneverse,yeshwanth2025excap3d0,Intent_3D} has focused on referring to objects using explicit names or attributes. 
Recent million-level extensions such as 3D-GRAND~\cite{yang2024_3D_GRAND} and SceneVerse~\cite{jia2024sceneverse} introduce richer annotations, including multi-object grounding. However, these data sets still mainly rely on straightforward object references rather than complex spatial relationships. Consequently, they allow models to exploit shortcut biases, limiting their ability to evaluate true spatial understanding.
Recently, ExCap3D~\cite{yeshwanth2025excap3d0} explores expressive captioning in 3D scenes at multiple levels of detail covering more than 3k objects. This work is inspiring because of the rich language descriptions for abundant object classes, but not directly focus on spatial segmentation or reasoning. Recent advancement like Intent3D~\cite{Intent_3D} introduce grounding based on human intention, but still rely on semantic cues for detecting object categories, ignoring the spatial relationship inherent in 3D. ScanReason expands this by proposing multiple reasoning types, including spatial and safety reasoning. However, its spatial component remains relatively coarse and lacks detailed supervision, such as segmentation masks.

\subsection{3D Large Language Model}

The advent of large language models (LLMs) has led to more sophisticated 3D-VL models emphasizing spatial and reasoning capabilities \cite{leo, jia2024sceneverse, li2024m3dbench}. For example, 3D-LLM injects 3D spatial knowledge into pre-trained LLMs, facilitating nuanced scene understanding. Models like LEO~\cite{leo} integrate language models with 3D understanding, enabling open-ended scene reasoning and interaction capabilities. Chat-3D~\cite{Chat-3D}, Chat-Scene~\cite{CHATSCENE}, and Grounded 3D-LLM~\cite{grounded3DLLM}, incorporate explicit identifiers or referent tokens to bridge linguistic queries and specific scene elements. Recent efforts explicitly focus on reasoning-guided segmentation tasks. For example, Reason3D~\cite{reason3d} and SegPoint~\cite{segpoint} employ LLM-driven frameworks to guide point cloud segmentation based on natural language instructions. MORE3D~\cite{jiang2024multimodal} and MLLM-For3D~\cite{MLLMfor3D} adopt multimodal LLMs (MLLMs) to simultaneously reason about complex spatial relations and output detailed segmentation masks.


%% file: sec/3_benchmark.tex
\section{3D Spatial Reasoning Segmentation (3D-SRS): Task and Benchmark}
\label{3drsr}

\subsection{Task Definition}
We define 3D Spatial Reasoning Segmentation (3D-SRS) as follows. Given a 3D scene $\mathcal{S}$ (e.g., a reconstructed indoor room) and a language query $q$ describing a spatial relation, the goal is to produce a segmentation mask $M$ that highlights all object(s) in $\mathcal{S}$ satisfying $q$. Formally, we learn a function $f(\mathcal{S},q)\to M$, where $M\subseteq\mathcal{S}$ consists of the 3D points (or object regions) referred to by the query. For example, if $q=$“The chair closest to the door”, then $M$ contains the points of the chair nearest the door. If multiple objects satisfy $q$, the mask covers all of them.

\begin{figure}[!t]
    \centering
    \includegraphics[width=\textwidth]{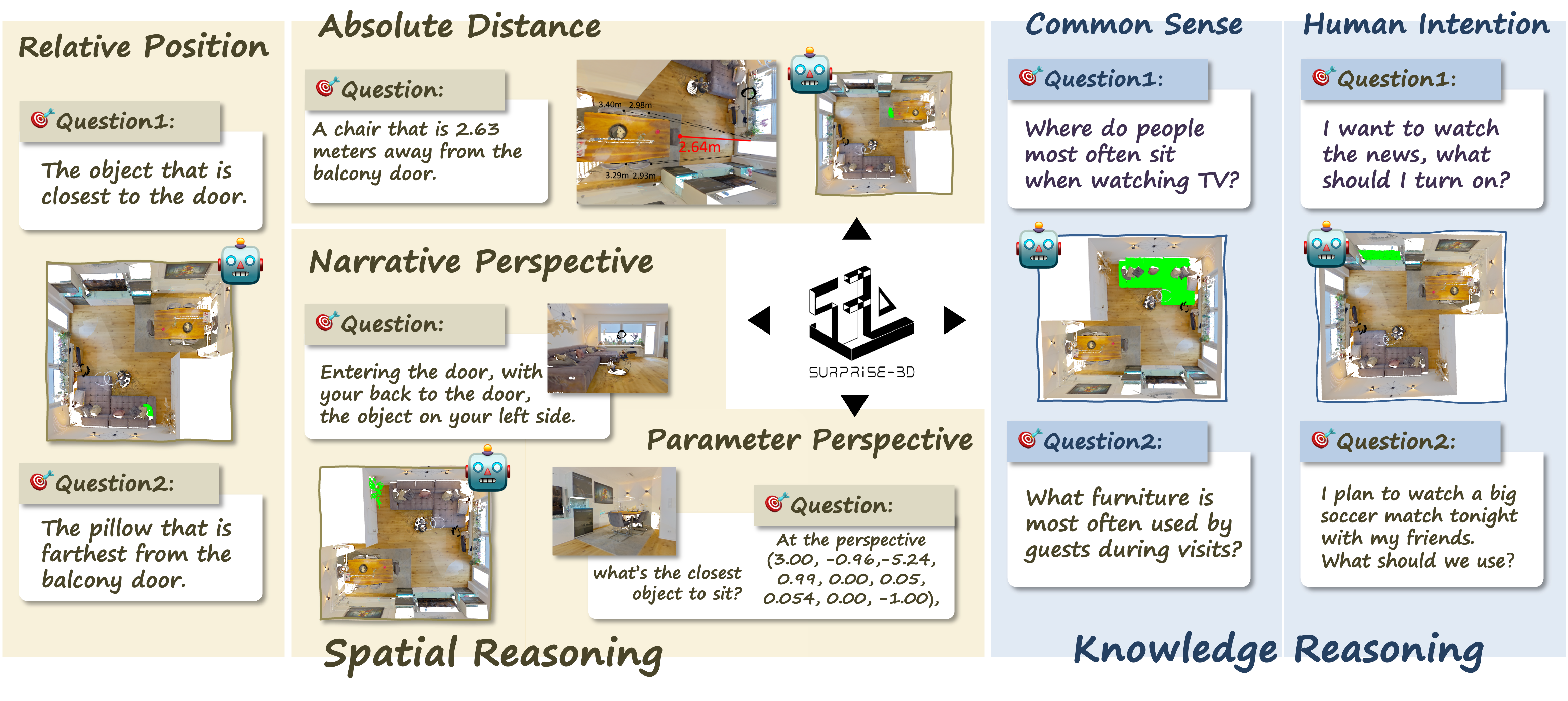}
    \caption{Examples of query categories in our 3D-SRS task.}
    \label{fig:task_define}
\vspace{-10 pt}
\end{figure}

\textbf{Narrative perspective}: interpreting egocentric references from a described point of view. For example, a query might implicitly place an agent in the scene (e.g., sitting on the black sofa facing the blackboard, the object to your left used for teaching). The model should simulate the narrator's point of view and understand terms like 'to your left' or 'in front of you' in this context.

\textbf{Parametric perspective}: understanding the current orientation and position parameters given in the description. For instance, \textit{facing away from the door towards the cabinet} specifies a camera orientation. The model must parse such instructions about where the observer is looking or located and use them to ground other spatial terms.

\textbf{Relative position}: reasoning about spatial relations between objects. Queries frequently use relational phrases such as "on the table", "behind the sofa", "to the left of the cabinet", etc. The model must identify reference objects (table, sofa, cabinet) and understand directional or topological relations (on, behind, left of) to find the target. This includes handling occlusion (e.g., an object "behind the sofa") that may be partially or fully hidden from certain viewpoints, requiring true 3D understanding beyond a single-camera view.

\textbf{Absolute Distance-based reasoning}: interpreting absolute or comparative distance cues. Many queries involve terms such as 'closest', 'furthest', or 'near'. The model should be able to compare distances between multiple objects with the same semantic category. For example, \textit{'the table closest to the bed'} requires finding all tables in the scene and selecting the one with minimal distance to the bed. Importantly, such distance terms are defined in an absolute spatial sense (the physical distances between objects), rather than relative to the viewer’s perspective. The model must therefore compute and compare 3D distances or understand spatial superlatives in context.

\begin{figure}[!t]
    \centering
    \includegraphics[width=\textwidth]{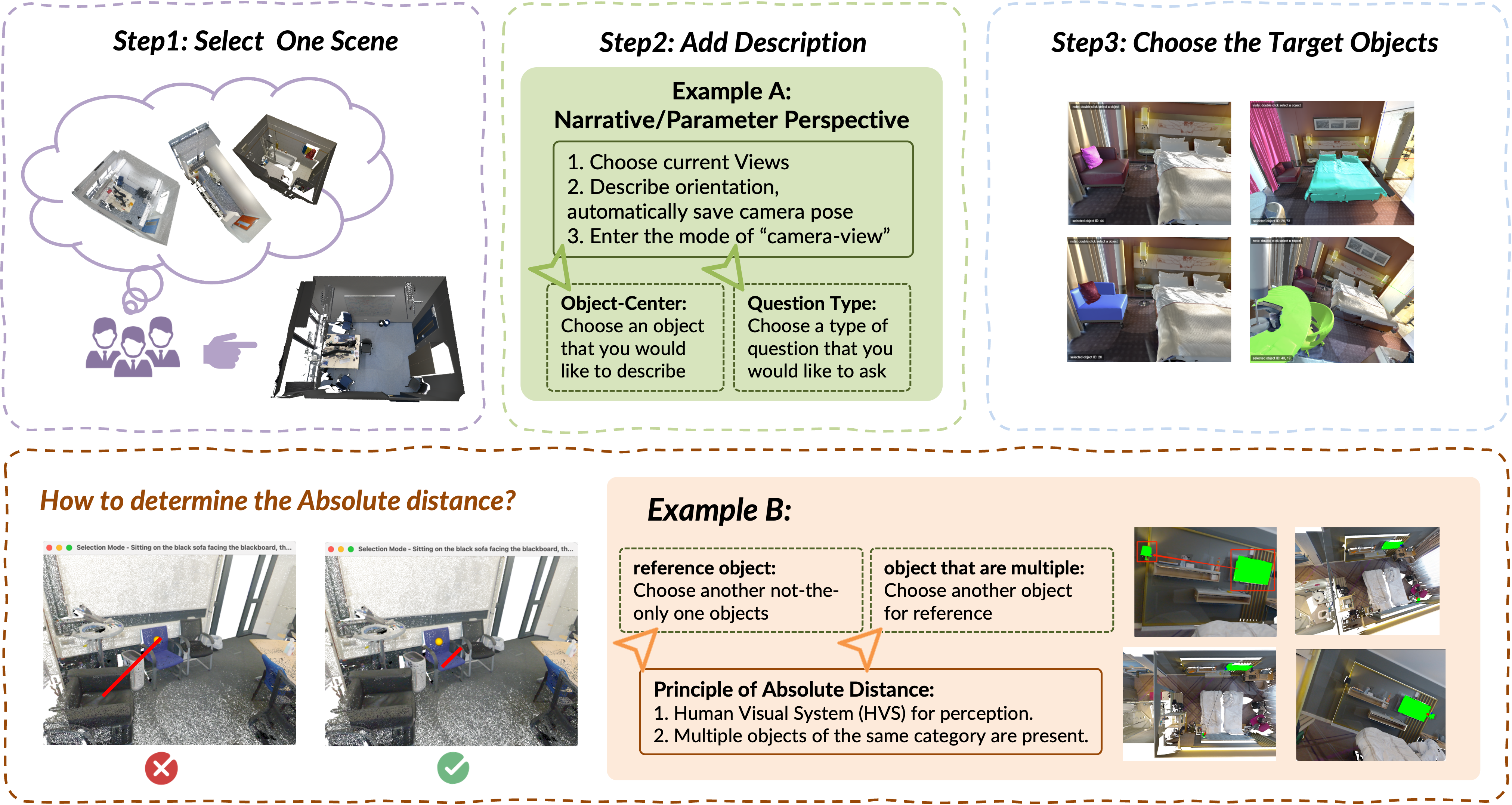}
    \caption{\textbf{Spatial reasoning annotation pipeline.} Human Annotators select a scene and target object, then write a question that identifies the object via spatial context, and finally mask the object.}
    \label{fig:spatial_anno}
    \vspace{-1.2em}
\end{figure}

\subsection{Benchmark Protocol and Evaluation}
We divide the data into disjoint training and validation splits at the scene level to prevent overlap.  For evaluation, we adopt standard segmentation metrics: the primary score is \emph{mean Intersection-over-Union} (mIoU) between the predicted and ground-truth masks across all queries.  We also report precision and recall at fixed IoU thresholds (e.g., 0.5 and 0.75) to analyze performance sensitivity. These metrics follow common practice in referring segmentation benchmarks~\cite{tgnn,3d-stmn,qian2024x}.  

To support a leaderboard, we release train/val annotations. Participants train models on the public data and submit predicted masks on the test scenes.  The benchmark server computes all metrics on the withheld ground truth to rank submissions. This protocol ensures fair comparison with identical data splits.

\section{S\textsc{urprise}3D: Dataset Construction and Annotation}

We constructed S\textsc{urprise}3D with two parallel annotation pipelines to capture spatial reasoning, common sense, and human intention reasoning. One pipeline focuses on spatial reasoning, where annotators formulate four type of questions and mask the corresponding target objects in 3D scenes. The other pipeline focuses on common sense and human intention queries, where the questions probe typical human knowledge or intent in the scene and require identifying the object that satisfies the query. By design, these pipelines operate independently but on the same set of scenes, ensuring a rich and complementary set of annotations. In total, S\textsc{urprise}3D provides a balanced mix of query types (spatial vs. knowledge-based) and a ground-truth target object for each query. In the following, we describe each annotation process and then analyze the coverage and balance of the dataset.


\subsection{Annotation Pipeline}
\begin{figure}[!t]
    \centering
    \includegraphics[width=\textwidth]{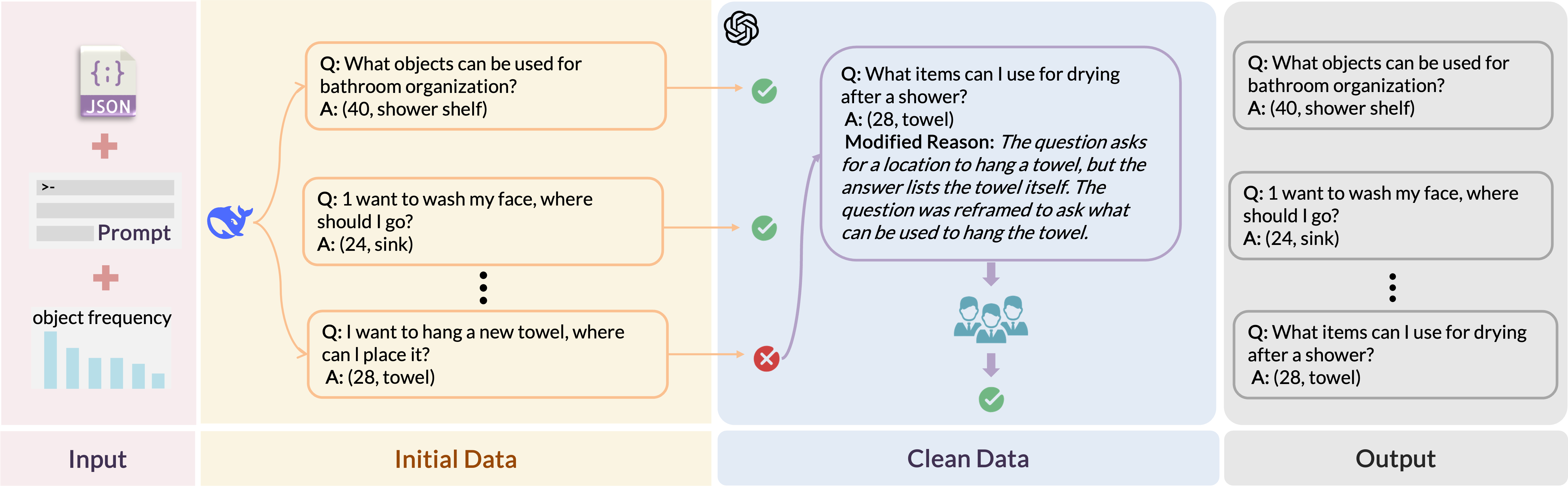}
    \caption{Overview of the common sense and human intention reasoning queries generation pipeline.}
    \label{fig:auto_pipe}
\end{figure}

We use separate annotation pipelines for spatial vs. knowledge queries (see Figure~\ref{fig:spatial_anno} and Figure~\ref{fig:auto_pipe}).

\textbf{Spatial reasoning.} Annotators view each 3D scene from a fixed camera viewpoint (e.g., top-down or entry view) and manually identify the object(s) satisfying the spatial query.  The interface allows for clicking or highlighting points in the rendered scene. For example, to answer 'closest to the door', the annotator selects the object closest to the entrance point of the door. This produces a ground-truth mask for the target object. Using a fixed viewpoint ensures that spatial relations can be assessed consistently from one perspective. 

First, annotators could freely navigate or choose a fixed camera viewpoint in the 3D scene. Note that the target objects can be visible or invisible from this camera view. 
Locking this point of view is critical, as it establishes a clear 'frame' of reference for egocentric directions such as left or right. The interface then allowed the annotator to enter a description of a target object based on that perspective, and finally to mark the object’s mask directly in the point cloud of 3D scenes. We recorded the camera parameters (extrinsics and intrinsics) alongside each query, so that any model processing the data can interpret the spatial language from the correct viewpoint context. The annotation UI thus lets the annotator 'be the agent' in the scene, selecting an orientation and then writing a query as if they were there (for example, an annotator might place the camera facing a wall and then describe 'the chair to my right' relative to that orientation). Once the description was written, the annotator highlighted the referenced object by drawing its segmentation mask over the 3D point cloud (the tool projected the mask onto the 3D object surface). This yields a ground-truth mask for the query.

\textbf{Common sense and human intention reasoning.} For common-sense and intention queries, we use an LLM-augmented pipeline inspired by recent work. We first generate candidate question-answer pairs using a large language model, given scene metadata and object labels as context. The LLM proposes questions (e.g., 'What objects can be used for bathroom organization? A: (shower shelf)'). These are automatically filtered by basic rules (removing duplicates or nonsensical queries). The remaining questions are then passed on to human annotators for validation. Figure~\ref{fig:auto_pipe} illustrates this two-stage process: LLM outputs are checked by humans, who either accept them or refine the question/answer. Invalid or ambiguous queries (marked with \xmark) are corrected or discarded, while valid ones (\checkmark) become part of the dataset. This workflow scales up annotation efficiently while retaining human oversight.

\subsection{Annotation Strategy and Quality Control}
\begin{figure}[!t]
    \centering
    \includegraphics[width=0.9\textwidth]{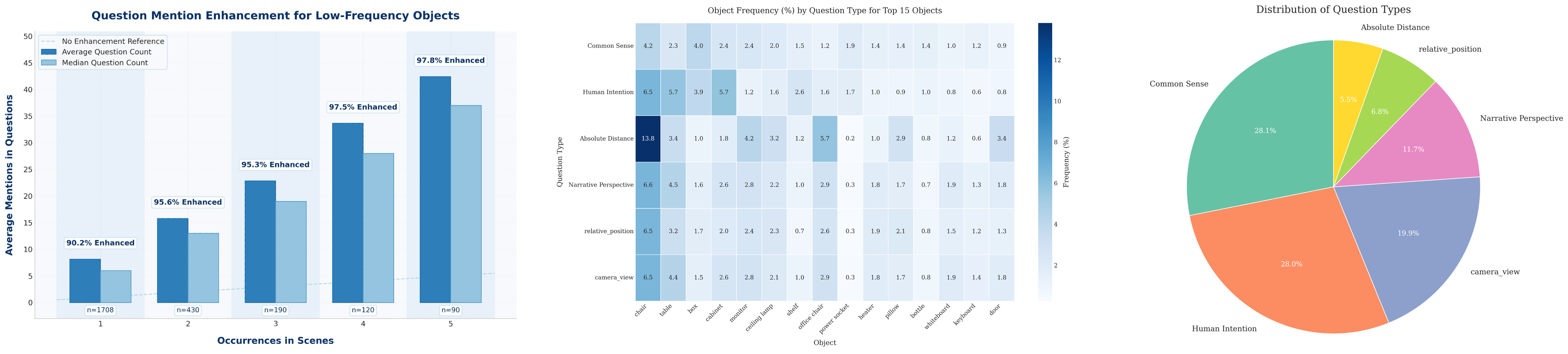}
    \caption{S\textsc{urprise}3D Dataset Statistics and Enhancements.}
    \label{fig:data_ana}
\end{figure}

To ensure clarity and coverage, we apply: 
(1) \textbf{Disambiguation Guidelines:} We require that the questions be unambiguous. Annotators are given rules (e.g., explicitly naming reference objects) to avoid confusion. Queries with multiple plausible answers are rephrased or omitted. 
(2) \textbf{Rare-Object Sampling:} To improve the representation of uncommon classes, we identify objects with low overall frequency and sample scenes containing them. We then generate additional queries targeting these rare objects. This rare object boost increases their mentions (on average by $\sim$90–100\%), improving the learning of these categories, as shown in Figure~\ref{fig:data_ana}. 
(3) \textbf{Human Verification:} All annotations are double-checked. Spatial queries receive multiple annotator checks for the target mask, and discrepancies are resolved by consensus. LLM-generated queries are reviewed by editors to ensure that the question matches the intended object. This multi-stage review yields high-quality, consistent annotations.

%% file: sec/4_experiment.tex
\section{Experiments}
In this section, we first provide a brief overview of the advanced methods evaluated on our proposed dataset, models setting including zero-shot and fine tuning are also introduced. We then describe the evaluation metrics and criteria used to assess the performance of these methods on knowledge and spatial reasoning tasks. Finally, we present the quantitative and qualitative results obtained by the aforementioned methods to comparatively analyze their performance on the newly introduced reasoning challenges.

\begin{table}[!t]
\setlength{\tabcolsep}{6.0pt} 
\fontsize{8.3}{8.5}\selectfont 
\caption{The results of various methods on different reasoning tasks under \textbf{zero-shot} setting.}
\label{table:2a}
\centering
\begin{tabular}{l c c c c c c c}
\toprule
\multirow{2}{*}{\vspace{-0.6em}Task Type} & \multirow{2}{*}{\vspace{-0.6em}Metric} & \multicolumn{6}{c}{Method}\\
\cmidrule(l{1pt}r{1pt}){3-8}
& & {MLLMfor3D} & {3D-Vista} & {Reason3D} & {Intent3D} & {ChatScene} & {Avarage}\\
\midrule
\multicolumn{8}{c}{\textbf{\small\textit{Knowledge Reasoning}}}\\
\midrule
\multirow{3}{*}{Common-sense}
& A${25}$ & \textbf{20.42} & 6.14 & \,\,\,6.97 & 10.01 & \,\,\,7.86 &10.28 \\
& A${50}$ & \textbf{13.42} & 6.14 & \,\,\,3.11 & 3.24 & \,\,\,4.01 &5.98\\
& mIoU & \textbf{12.75} & -- & \,\,\,4.79 & -- & -- &8.77\\
\cmidrule(l{0.5pt}r{0.5pt}){1-8}
\multirow{3}{*}{Human Intention}
& A${25}$ & \textbf{22.38} & 8.26 & 11.33 & 15.84 & \,\,\,1.64& 11.89 \\
& A${50}$ & \textbf{13.62} & 8.26 & \,\,\,6.03 & 5.82 & \,\,\,1.00 &6.95\\
& mIoU & \textbf{11.91} & -- & \,\,\,7.51 & -- & -- &9.71 \\
\cmidrule(l{0.5pt}r{0.5pt}){1-8}

\multicolumn{8}{c}{\textbf{\small\textit{Spatial Reasoning}}}\\
\midrule
\multirow{3}{*}{Narrative Perspective}
& A${25}$ & \textbf{15.07} & 6.50 & 8.39 & 2.65 & \,\,\,0.00 &6.52\\
& A${50}$ & \textbf{13.62} & 6.50 & 4.88 & 0.77 & \,\,\,0.00 &5.15 \\
& mIoU & \textbf{11.56} & -- & 5.63 & -- & --&8.60 \\
\cmidrule(l{0.5pt}r{0.5pt}){1-8}
\multirow{3}{*}{Parametric Perspective}
& A${25}$ & 4.25 & 3.65 & \textbf{7.91} & 2.62 & \,\,\,0.00&3.69 \\
& A${50}$ & 3.20 & 3.65 & \textbf{4.82} & 0.79 & \,\,\,0.00&3.12 \\
& mIoU & 2.93 & -- & \textbf{5.68} & -- & --&4.31 \\
\cmidrule(l{0.5pt}r{0.5pt}){1-8}
\multirow{3}{*}{Relative Position}
& A${25}$ & 7.78 & 6.52 & \textbf{9.57} & 4.30 & \,\,\,1.38&5.91 \\
& A${50}$ & 5.81 & \textbf{6.52} & 6.38 & 0.97 & \,\,\,0.00 &3.94\\
& mIoU & 4.92 & -- & \textbf{6.78} & -- & -- &5.85\\
\cmidrule(l{0.5pt}r{0.5pt}){1-8}
\multirow{3}{*}{Absolute Distance}
& A${25}$ & \textbf{11.90} & 7.61 & 9.10 & 2.41 & \,\,\,1.39 &6.48\\
& A${50}$ & \textbf{10.62} & 7.61 & 2.60 & 0.74 & \,\,\,0.00 &4.32\\
& mIoU & \textbf{9.24} & -- & 5.25 & -- & -- &7.25\\
\midrule
\multicolumn{8}{c}{\textbf{\small\textit{Overall Reasoning}}}\\
\midrule
\multirow{3}{*}{{Overall}}
& A${25}$ & \textbf{13.63} & 6.40 & 9.09 & 8.63 & \,\,\,3.59 &8.27\\
& A${50}$ & \textbf{10.05} & 6.40 & 4.57 & 2.94 & \,\,\,1.77 &5.15\\
& mIoU & \textbf{8.89} & -- & 6.08 & -- & -- &7.49\\
\bottomrule
\end{tabular}
\end{table}

\subsection{Baselines}
Spatial reasoning tasks require a deeper understanding of semantic relationships and spatial configurations within 3D scenes. To evaluate the effectiveness of existing approaches on these tasks, we conduct comprehensive experiments on several advanced methods, including MLLMfor3D \cite{MLLMfor3D}, 3D-Vista \cite{3dvista}, Reason3D \cite{chen2024reasoning3d}, ChatScene \cite{CHATSCENE} and Intent3D \cite{Intent_3D}. All these methods take natural language questions or descriptions as input, and output masks or bounding boxes of objects that are the answer of given prompts in 3D scenes. MLLMfor3D \cite{MLLMfor3D} is a label-free paradigm of 3D understanding, it projects 2D pseudo-masks of objects to 3D scene. 3D-Vista \cite{3dvista} directly encourages the alignment of masked text with masked 3D scenes. Reason3D \cite{chen2024reasoning3d} utilizes a pre-trained LLM to process input point and text features, and predict segmentation masks of targeted objects. Intent3D \cite{Intent_3D} aligns the point features of scenes and candidate box features integrated with the encoded prompts.
\subsection{Model settings}
\textbf{Zero-shot}.
To assess the capabilities of existing methods in performing knowledge-based and spatial reasoning tasks, we directly evaluate pre-trained models, including MLLMfor3D \cite{MLLMfor3D}, 3D-Vista \cite{3dvista}, Reason3D \cite{reason3d}, ChatScene \cite{CHATSCENE} and Intent3D \cite{Intent_3D} on our proposed dataset, without training or finetuning on our datasets. The inputs consist of 3D scenes represented as point clouds, paired with images and textual questions that focus on knowledge and spatial reasoning. As reported in previous work \cite{MLLMfor3D, reason3d, Intent_3D, 3dvista, CHATSCENE}, Intent3D, 3D-Vista, ChatScene and Reason3D demonstrate partial abilities to understand human intentions and common sense knowledge. Moreover, MLLMfor3D and 3D-Vista show promising results in reasoning about relative object positions. However, none of them has been evaluated or shown to have the ability to comprehend narrative perspectives, parametric perspectives, or absolute spatial distances. Our dataset serves as the first benchmark to systematically evaluate these unexplored dimensions of spatial and knowledge understanding in existing 3D vision language models.

\textbf{Fine tuning}.To explore the ability of current advanced methods to learn knowledge and spatial reasoning, we finetune MLLMfor3D, 3D-Vista, Reason3D, and Intent3D on our dataset. 

\begin{table}[!t]
\setlength{\tabcolsep}{6.0pt}  
\fontsize{8.3}{8.5}\selectfont 
\caption{The results of various methods on different reasoning tasks under \textbf{fine-tuned} setting. }
\label{table:2b}
\centering
\begin{tabular}{l c c c c c c c}
\toprule
\multirow{2}{*}{\vspace{-0.6em}Task Type} & \multirow{2}{*}{\vspace{-0.6em}Metric} & \multicolumn{6}{c}{Method}\\
\cmidrule(l{1pt}r{1pt}){3-8}
& & {MLLMfor3D} & {3D-Vista} & {Reason3D} & {Intent3D} & {ChatScene} & {Avarage}\\
\midrule
\multicolumn{8}{c}{\textbf{\small\textit{Knowledge Reasoning}}}\\
\midrule
\multirow{3}{*}{Common-sense}
& A${25}$ & 25.41 & 19.36 & 18.08 & \textbf{30.09} & 13.56 &21.30 \\
& A${50}$ & \textbf{23.92} & 19.36 & 8.97 & 15.22 & \,\,\,4.37 &14.37\\
& mIoU & \textbf{19.40} & -- & 11.92 & -- & -- &15.66\\
\cmidrule(l{0.5pt}r{0.5pt}){1-8}
\multirow{3}{*}{Human Intention}
& A${25}$ & 28.42 & 22.36 & 17.98 & \textbf{31.16} & 13.80 &22.74\\
& A${50}$ & \textbf{24.51} & 22.36 & 10.81 & 18.08 & \,\,\,4.70 & 16.09\\
& mIoU & \textbf{19.47} & -- & 11.98 & -- & -- &15.73\\
\cmidrule(l{0.5pt}r{0.5pt}){1-8}

\multicolumn{8}{c}{\textbf{\small\textit{Spatial Reasoning}}}\\
\midrule
\multirow{3}{*}{Narrative Perspective}
& A${25}$ & 22.38 & \textbf{25.77} & 11.30 & 24.91 & 13.98 &19.67\\
& A${50}$ & 20.40 & \textbf{25.77} & 7.71 & 15.58 & \,\,\,4.28 &14.75\\
& mIoU & 18.44 & -- & 9.03 & -- & -- &13.74\\
\cmidrule(l{0.5pt}r{0.5pt}){1-8}
\multirow{3}{*}{Parametric Perspective}
& A${25}$ & 10.04 & 3.87 & 11.52 & \textbf{12.57} & \,\,\,4.58 & 8.52\\
& A${50}$ & \textbf{9.33} & 3.87 & 7.59 & 6.13 & \,\,\,2.36 &5.86\\
& mIoU & 7.50 & -- & \textbf{8.85} & -- & -- &8.18\\
\cmidrule(l{0.5pt}r{0.5pt}){1-8}
\multirow{3}{*}{Relative Position}
& A${25}$ & 22.61 & \textbf{23.86} & 11.51 & 12.90 & 12.48 &16.67\\
& A${50}$ & 18.76 & \textbf{23.86} & 8.18 & 7.49 & \,\,\,1.39 &11.94\\
& mIoU & 14.70 & -- & 8.88 & -- & -- &11.79\\
\cmidrule(l{0.5pt}r{0.5pt}){1-8}
\multirow{3}{*}{Absolute Distance}
& A${25}$ & \textbf{25.30} & 18.92 & 12.80 & 5.75 & \,\,\,7.42 &14.04\\
& A${50}$ & \textbf{20.37} & 18.92 & 4.45 & 1.86 & \,\,\,3.71 &9.86\\
& mIoU & \textbf{19.23} & -- & 8.16 & -- & -- &13.70\\

\midrule
\multicolumn{8}{c}{\textbf{\small\textit{Overall Reasoning}}}\\
\midrule
\multirow{3}{*}{Overall}
& A${25}$ & 22.36 & 18.34 & 16.14 & \textbf{23.98} & 11.58 &18.48\\
& A${50}$ & 19.55 & 18.34 & 9.06 & \textbf{13.10} & \,\,\,3.99 &12.81\\
& mIoU & 16.46 & -- & 11.00 & -- & -- &13.73\\
\bottomrule
\end{tabular}
\end{table}

\subsection{Evaluation metrics}
For the segmentation tasks (MLLMfor3D and Reason3D), we adopt both Mean Intersection over Union (MIoU) \cite{MIoU} and Accuracy (Acc) as evaluation metrics. MIoU measures the average overlap between the predicted and true 3D volumes, while Accuracy evaluates precision across varying confidence thresholds, which we obtain from different intersection proportions (e.g., 0.25 and 0.50) of the predicted and ground-truth volumes. For the detection tasks (3D-Vista, Intent3D, and ChatScene), we use Accuracy as the sole evaluation metric.
\subsection{Results analysis}
\begin{figure}[!t]
    \centering
    \includegraphics[width=1.0\textwidth]{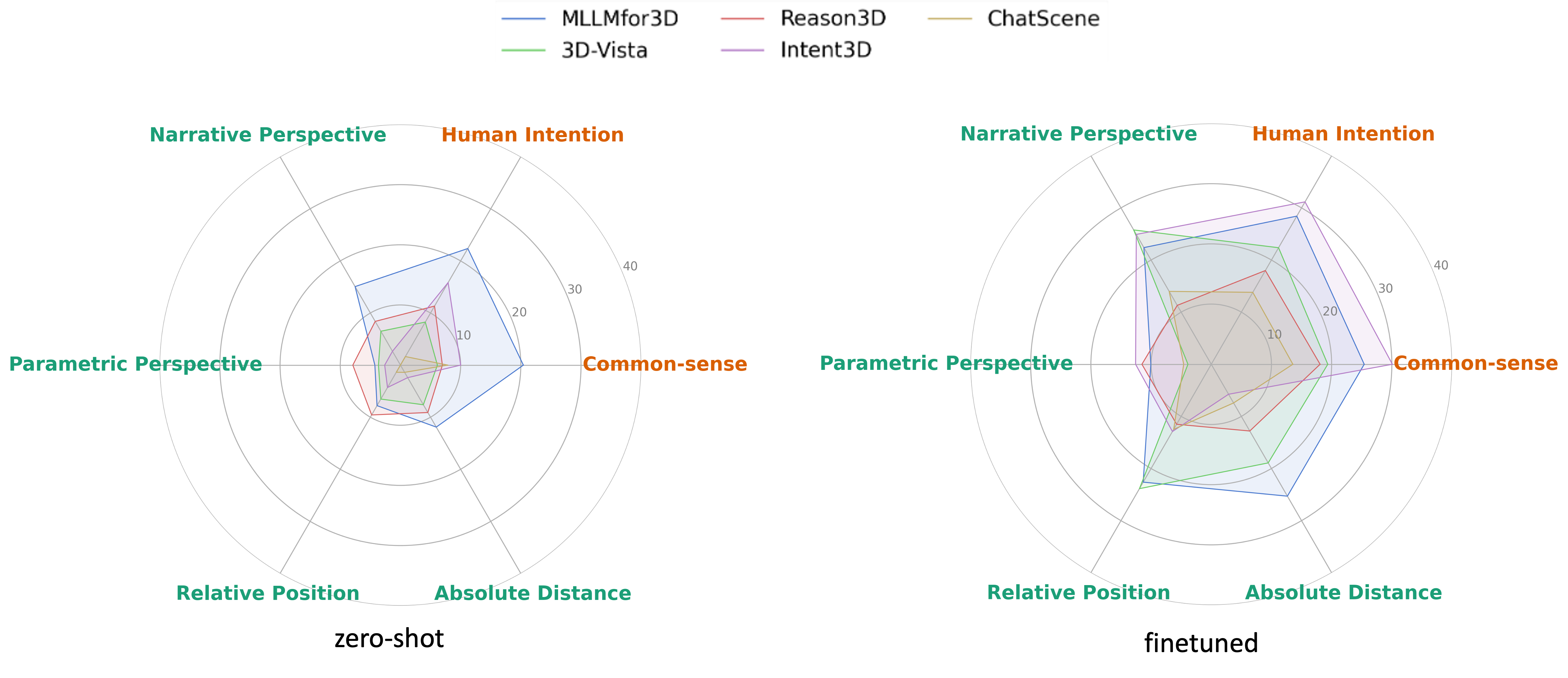}
    \caption{The comparison between zero-shot and fine-tuned models on all reasoning tasks.}
    \label{fig:radar_chart}
\end{figure}

As shown in Table~\ref {table:2a}, the zero-shot results indicate that all models demonstrate relatively weak overall spatial reasoning capabilities compared with knowledge reasoning capabilities. Although these methods have been trained on spatial description datasets, such as ScanRefer~\cite{chen2020scanrefer} for most models and SQA3D~\cite{ma2022sqa3d} for ChatScene. After fine-tuning on our dataset, as presented in Tables~\ref{table:2b} and Figure~\ref{fig:radar_chart}, all models demonstrate substantial improvements in reasoning abilities. This improvement is particularly significant in spatial reasoning, with an average performance increase of approximately three times. We hypothesize that this improvement is due to a key difference between the existing data sets and our proposed dataset. Although prior data sets contain questions involving spatial information, they often retain excessive semantic cues that can serve as shortcuts for models to arrive at answers without fully exercising spatial reasoning. In contrast, our dataset deliberately removes such shortcuts, ensuring that the training process emphasizes and encourages the spatial reasoning capability of the models. These findings highlight that current methods still have substantial room for improvement in terms of spatial reasoning, revealing an important avenue for future research.

\label{resultanalysis}

\section{Conclusion and Limitations}

We introduced \textbf{S\textsc{urprise}3D}, a large-scale dataset and benchmark for evaluating spatial and knowledge reasoning in 3D scenes. Our benchmark defines the 3D Spatial Reasoning Segmentation (3D-SRS) task, which includes various types of spatial queries, including relative position, absolute distance, narrative perspective, and parametric viewpoint, as well as commonsense and human intention grounding. Through a dual annotation pipeline and rare-object enhancement, S\textsc{urprise}3D provides high-quality, diverse, and spatially language-query and segmentation pairs. Extensive analysis confirms its balanced coverage and strong potential for advancing spatial intelligence in 3D vision-language models.

\textbf{Limitations.} While our human annotation ensures quality, it limits scalability. Some query types (e.g., parametric view) may be less natural for real-world deployment. In addition, annotations are restricted to indoor scenes from ScanNet++, which may not generalize to outdoor or dynamic environments. We leave domain transfer, temporal reasoning, and multi-turn interaction as future directions.